%% file: main.tex
\documentclass[letterpaper,10pt,conference]{ieeeconf}
\IEEEoverridecommandlockouts
\usepackage{amsmath,amssymb,array,booktabs,cite,colortbl,graphicx,multirow,needspace,placeins,stfloats,tabularx,xcolor}

\makeatletter
\newenvironment{LayoutInlineTable}{\begingroup\footnotesize\def\@captype{table}}{\endgroup}
\makeatother

\newcommand{\method}{\textsc{PredTac}}
\newcolumntype{Y}{>{\centering\arraybackslash}X}
\newcolumntype{L}[1]{>{\raggedright\arraybackslash}p{#1}}
\newcolumntype{C}[1]{>{\centering\arraybackslash}p{#1}}
\newcolumntype{Z}{>{\raggedright\arraybackslash}X}
\definecolor{predblue}{RGB}{44,105,170}
\definecolor{touchorange}{RGB}{230,126,34}
\definecolor{ctrlgreen}{RGB}{60,145,90}
\definecolor{softgray}{RGB}{245,247,249}

\providecommand{\RealMethodExtension}{}
\providecommand{\RealExperimentExtension}{}
\providecommand{\RealResultsExtension}{}
\providecommand{\RealResultsDisplay}{}
\providecommand{\RealDiscussionBoundary}{All control results are simulation-only,
goal-offset diagnostics conditional on fixed checkpoints. P/P
and Vision each use one policy-training seed, whereas T/T training-seed
provenance is incomplete; these comparisons do not assess retraining variability.
Broad P/0$_b$ also removes goal touch and predictor execution, while Vision
uses a separately trained controller without tactile inputs.
The study therefore does not establish standard-reset full-task generalization
or real-robot performance.}
\providecommand{\RealDiscussionExtension}{}

\providecommand{\RealConclusionExtension}{}

\input{real_extension_en.tex}

\title{\LARGE \bf PredTac: Learning Contact-Rich Manipulation\\
with Predicted Touch}
\author{Weijia Fan and Daqiang Guo\textsuperscript{*}%
\thanks{\raggedright The Hong Kong University of Science and
Technology (Guangzhou), Guangzhou, China.
\newline Emails: wfan768@connect.hkust-gz.edu.cn;
daqiangguo@hkust-gz.edu.cn.
\newline
\textsuperscript{*}Corresponding author: Daqiang Guo.}}

\begin{document}
\bstctlcite{BSTcontrol:etal}
\maketitle
\thispagestyle{empty}
\pagestyle{empty}

\begin{abstract}
Contact-rich manipulation benefits from tactile feedback, yet physical tactile
sensors introduce hardware, calibration, synchronization, and maintenance
costs that complicate policy learning and deployment. We formulate predicted
touch as an alternative to measured tactile input and present PredTac, a
framework that learns to infer tactile states from causal visual observations
and robot states and uses the predicted touch as an explicit interface for
policy learning and execution. A tactile predictor is first trained with
tactile supervision and then used to provide contact information without
requiring measured tactile input during downstream policy training or
execution. We evaluate PredTac across three contact-rich manipulation tasks
in simulation and on a real robot, and further examine how policy performance
depends on the predicted contact content. In simulation goal-offset
evaluations, predicted-touch policies achieve 27.0\%, 52.0\%, and 44.7\%
success on USB, Barbed-spike, and Valve, respectively, improving over the
visual baseline by 8.0--13.7 percentage points. On the real robot,
predicted-touch ACT achieves 70.0\%, 50.0\%, and 90.0\% success on USB
insertion, Barbed extraction, and Valve rotation, respectively, with a
three-task mean of 70.0\%, approaching measured-touch ACT at 72.2\% and
substantially outperforming visual ACT at 21.1\%. Fixed-policy interventions
further show that performance is sensitive to the spatial structure of
predicted contact, with spatial rearrangement at fixed value distributions
reducing Valve success by 10.7 percentage points. These results demonstrate
that predicted touch can provide useful contact information for contact-rich
manipulation without requiring tactile sensing as a policy input.
\end{abstract}

\section{Introduction}
Contact-rich manipulation is only partially observable through cameras, as
visually similar states can correspond to very different contact conditions.
For example, during USB insertion, the same wrist view may indicate no
contact, edge contact, or partial seating, while subtle changes in alignment,
local support, pressure distribution, or release can determine success yet
remain visually ambiguous. Tactile sensing exposes these events, and visuotactile policies can
benefit from access to contact states~\cite{luu2025manifeel}. Physical sensors,
however, introduce calibration, synchronization, and maintenance requirements.
We ask whether a learned touch predictor can support policy training and
execution without measured touch as a policy input, while retaining useful
contact information.

Prior work learns visual--tactile mappings
~\cite{lee2019touching,li2019connecting,ayad2024imagine2touch,wu2025vitacgen}
or uses tactile supervision to train policies that omit measured touch at
inference~\cite{george2024vital,zhao2026fdvla,gubernatorov2026hapticvla}.
TacImag trains policies with observations from a frozen tactile
predictor~\cite{zhang2026tacimag}. These studies motivate learning contact
information from observations available at deployment. Our focus is the
explicit predicted-touch input and its role in downstream control.

Our study asks how well policies learn and act with predicted touch and which
aspects of its contact content affect control. Reconstruction error alone
cannot establish either the benefit of this input or how the policy uses it.

\begin{figure*}[t]
  \centering
  \includegraphics[width=\textwidth]{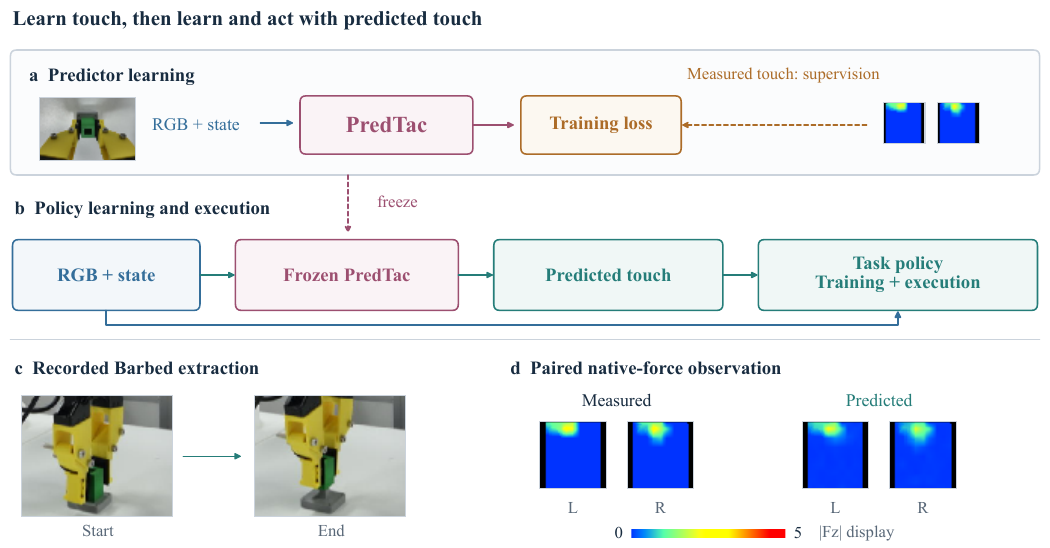}
  \caption{\textbf{PredTac supplies predicted touch during policy learning and execution.}
  (a) Measured touch supervises predictor learning from wrist RGB and robot-state
  histories. (b) The predictor is then frozen; the task policy trains and executes
  with predicted touch and current RGB/state. (c) Recorded start/end frames of
  Barbed extraction. (d) Paired measured and predicted $|F_z|$ for left (L) and
  right (R) fingers at one recorded Barbed observation. Maps share the runtime
  spatial projection and fixed 0--5 vendor-decoded display range; native signals
  are tri-axial taxel forces.}
  \label{fig:tasks}
\end{figure*}

\method{} uses a two-stage workflow.
Tactile supervision trains the predictor, which is then frozen for policy
learning and execution. This exposes the policy to the same predictor's
imperfect outputs at both stages. The simulation
implementation estimates current touch from three wrist-RGB frames and
matched states using spatiotemporal fusion, contact conditioning, and
bounded-residual decoding. The predicted field enters a ContactWorld
controller~\cite{zhang2026contactworld}. On the real robot, task-specific
predictors supply native dual-finger taxel forces to ACT policies alongside
current RGB and state. Figure~\ref{fig:tasks} illustrates this control
interface and recorded tactile predictions.

We compare predicted-touch, measured-touch, and visual controllers in simulation,
and test the role of current tactile content through input removal and
replacement. Auxiliary adaptation and multi-task studies examine reconstruction
and the use of different predictors with fixed policies. On the real robot,
we compare the three ACT conditions on USB, Barbed, and Valve, with 30 trials
per task and condition.

The contributions are:
\begin{itemize}
  \item We formulate predicted touch as a new paradigm for contact-rich
  manipulation, asking whether learned contact information can substitute
  for measured tactile input during policy learning and execution.
  \item We introduce PredTac, a unified framework for learning and exploiting
  predicted touch, enabling contact-rich manipulation through learned tactile
  representations derived from visual/state observations.
  \item We demonstrate the effectiveness of PredTac through simulation and
  real-robot experiments across three contact-rich tasks, achieving
  8.0--13.7 percentage-point gains over the visual baseline in simulation
  and a 70.0\% mean success rate on the real robot, compared with 72.2\%
  for measured-touch ACT and 21.1\% for visual ACT. We further evaluate
  cross-task predictor reuse on the simulated Valve rotation task. Keeping
  the policy fixed, replacing its original predictor with a shared predictor
  yields 44.0\% success, compared with 44.7\% before replacement. The shared
  predictor is fine-tuned on six other simulation tasks without Valve tactile
  data.
\end{itemize}

\section{Related Work}
\textbf{Cross-modal tactile prediction.}
Prior work learned visual--tactile correspondence~\cite{lee2019touching,li2019connecting},
predicted touch from geometry or depth patches~\cite{zapata2021generation,roefer2025pseudotouch,ayad2024imagine2touch},
or localized contact from RGB-D~\cite{kim2023im2contact}. Control-facing work
includes ViTacGen tactile images~\cite{wu2025vitacgen}, FELT dual-finger maps or
features~\cite{li2026felt}, TacImag force fields and imagined-observation
policies~\cite{zhang2026tacimag}, and TacGen scalable tactile latents with
matched-capacity evaluation~\cite{ye2026tacgen}.

\textbf{Touch-supervised deployment.}
VITaL reports non-tactile policies after visuotactile pretraining
~\cite{george2024vital}, while Ferrandis \emph{et al.} estimate state for control
under occlusion~\cite{ferrandis2025visuotactile}. FD-VLA distills force-aligned
tokens~\cite{zhao2026fdvla}; HapticVLA distills a tactile action expert into a
vision--state student~\cite{gubernatorov2026hapticvla}; NoContactNoWorries
predicts fingertip contact~\cite{patil2026nocontact}; and UniTac predicts tactile
representations without real-time touch~\cite{tu2026unitac}. ReTouch instead
refines predictions with execution-time tactile feedback~\cite{zhang2026retouch}.

\textbf{Visuotactile control and world models.}
Tactile MPC and action-conditioned predictors forecast touch for planning or
slip prediction~\cite{tian2019manipulation,mandil2022actp}; recent world models
forecast contact or joint visual--tactile futures~\cite{higuera2026vtwm,zang2026tacforesight,huang2026vitacworld,tian2026vtwam,ma2026feelworld}.
ManiFeel studies modality, representation, and policy design
~\cite{luu2025manifeel}; ContactWorld compares spatiotemporal representations
~\cite{zhang2026contactworld}; and TouchWorld combines tactile subgoals with
online feedback~\cite{zhou2026touchworld}.

\textbf{Positioning.}
TacImag and FELT use generated touch or features for downstream
policies~\cite{zhang2026tacimag,li2026felt}. \method{} combines task-policy
comparisons with fixed-policy tactile-input interventions in simulation
and evaluates native force predictions with real-robot ACT. These studies
distinguish task performance from dependence on predicted tactile content;
adaptation experiments examine input reuse.

\section{\method}
\begin{figure*}[t]
  \centering
  \includegraphics[width=\textwidth]{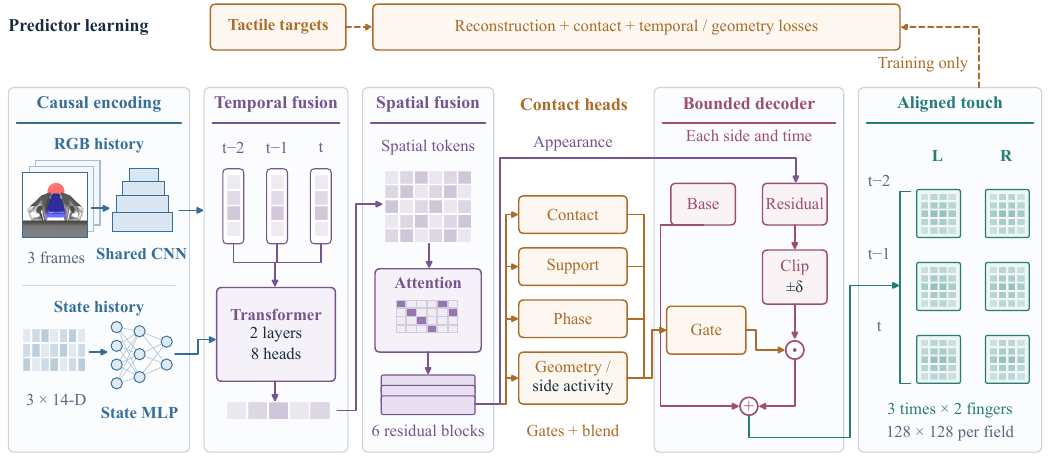}
  \caption{\textbf{PredTac predicts aligned tactile fields from visual and state histories.}
  Shared visual and state encoders feed temporal fusion and a spatial trunk.
  Appearance features and structured contact gates guide base-plus-bounded-residual
  decoding for each finger and observed time. Dashed arrows denote predictor-training
  supervision. The six outputs align with the three input times; they are not future
  forecasts. Feature tokens and field grids are schematic.}
  \label{fig:architecture}
\end{figure*}

\method{} supplies predicted touch to the policy alongside visual and state
inputs. The predictor is trained with tactile supervision and remains frozen
during policy learning and execution. Figure~\ref{fig:architecture} details
the tactile-field predictor.

\subsection{Problem formulation}
Let $v_t$ denote wrist RGB, $p_t\in\mathbb{R}^{14}$ the robot-state vector,
$\tau_t$ the tactile target, and $u$ an optional source/task context used by
a given predictor checkpoint. \method{} predicts
\begin{equation}
  \hat{\tau}_t=g_\phi(v_{t-2:t},p_{t-2:t},u).
  \label{eq:predictor}
\end{equation}
Equation~\eqref{eq:predictor} is causal: it uses only $t-2{:}t$. At deployment,
$g_\phi$ reads no measured touch/force, future observation, demonstration action,
or control outcome; \emph{predicted} denotes inference without runtime touch, not
$t+1$ forecasting.

The generic output is a normalized bilateral tactile field. In the ContactWorld
adaptation it is mapped to the simulator's right-side tactile-depth interface.
For the source pretrained checkpoint it represents a processed OmniViTac
reference-difference proxy, not calibrated depth or force. These targets have
different physical meanings, so their reconstruction scores are not directly
comparable.

\subsection{Spatiotemporal vision/state fusion}
Each RGB frame passes through a shared convolutional spatial encoder. The
robot state at the same timestamp is embedded by a
$14\!\rightarrow\!128\!\rightarrow\!256$ multilayer perceptron and fused with
256-D visual tokens. Two eight-head temporal Transformer layers aggregate the
two past tokens and the current token at each spatial site. Spatial
cross-attention and an additional encoder produce a 512-channel latent,
followed by six residual blocks. Because the input window ends at $t$, the
attention layers are non-anticipatory without a future mask.

\subsection{Structured contact conditioning and decoder}
The fused latent predicts contact state, support, interaction phase, side
activity, side area, and contact-shape descriptors. These outputs form a
structured branch that controls decoder gates and blends. Spatial features
also retain a direct appearance path to the decoder.

\Needspace{4\baselineskip}
A base-plus-residual decoder produces the side-specific field
\begin{equation}
  \hat{\tau}_{s,t}=b_s+
  \gamma(c_t,q_t,h_t)\odot
  \operatorname{clip}\!\left(\Delta\tau_{s,t},-\delta,\delta\right),
  \label{eq:decoder}
\end{equation}
where $s\in\{L,R\}$ indexes the tactile side, $b_s$ is a stable side-specific
base, $\Delta\tau_{s,t}$ is the learned appearance residual, $\delta$ bounds its
magnitude, and $\odot$ denotes elementwise multiplication. The gate $\gamma$ is
conditioned on contact-, phase-, and shape-related variables $(c_t,q_t,h_t)$.
Equation~\eqref{eq:decoder} is a schematic summary: the implementation also
contains support gating, release decay, and a conditioned base/residual blend.

The native generator emits three timestamps for two sides at
$128\!\times\!128$, ordered as
$[L_{t-2},R_{t-2},L_{t-1},R_{t-1},L_t,R_t]$. The ContactWorld runtime selects
the used side, resizes it to $320\!\times\!240$, and applies the policy's
$1\!\times\!80\!\times\!60$ preprocessing. Thus $320\!\times\!240$ is an
adapter resolution, not the native generator output.

\subsection{Learning objective}
Tactile targets supervise generation during training. The adaptation objective
combines full-field reconstruction, active-contact emphasis, temporal
difference consistency, contact classification, and checkpoint-dependent
geometry terms:
\begin{equation}
\begin{aligned}
  \mathcal{L}_{\rm touch}={}&\mathcal{L}_{\rm pix}
  +\lambda_a\mathcal{L}_{\rm active}
  +\lambda_\Delta\mathcal{L}_{\Delta}
  +\lambda_c\mathcal{L}_{\rm contact} \\
  &+\lambda_A\mathcal{L}_{\rm area}
  +\lambda_m\mathcal{L}_{\rm amp}
  +\lambda_\mu\mathcal{L}_{\rm centroid}.
\end{aligned}
  \label{eq:loss}
\end{equation}
In Eq.~\eqref{eq:loss}, $\mathcal{L}_{\rm pix}$ reconstructs the full field,
$\mathcal{L}_{\rm active}$ upweights active contact pixels,
$\mathcal{L}_{\Delta}$ penalizes temporal field differences,
$\mathcal{L}_{\rm contact}$ is computed from field-derived contact, and the
remaining terms match field-derived area, amplitude, and centroid. The
$\lambda$ coefficients are checkpoint-specific. The source checkpoint,
\emph{Pretrained}, was learned from 1,022 real-operation episodes (about 32.7k
frames), with tactile-derived proxy supervision for the structured heads.
ContactWorld target training supervises the final field and its statistics,
rather than those internal heads directly.

\subsection{Predictor training and reuse}
Target tactile supervision supports either fine-tuning the source checkpoint
or training the same architecture from scratch. The objectives use
active-weighted reconstruction and contact loss, optionally augmented by the
temporal and geometry terms in Eq.~\eqref{eq:loss}. USB training uses 161
episodes (10,892 frames), 20 validation episodes, 1,000 updates, and seed 42;
checkpoints are selected by validation MAE and then frozen.

For cross-task reuse, one right-touch \emph{shared predictor} is fine-tuned
with equal sampling weights for Barbed-flat, Barbed-spike, Lidded-loose, Peg,
Power, and USB. It uses active-weighted reconstruction and contact loss,
3,000 updates, and one training seed. Its source/task identifier remains fixed
to ``unknown'', so the same weights receive no task-varying token. Valve
contributes no tactile supervision to this predictor.

\begin{figure*}[t]
  \vspace{3pt}
  \centering
  \includegraphics[width=\textwidth]{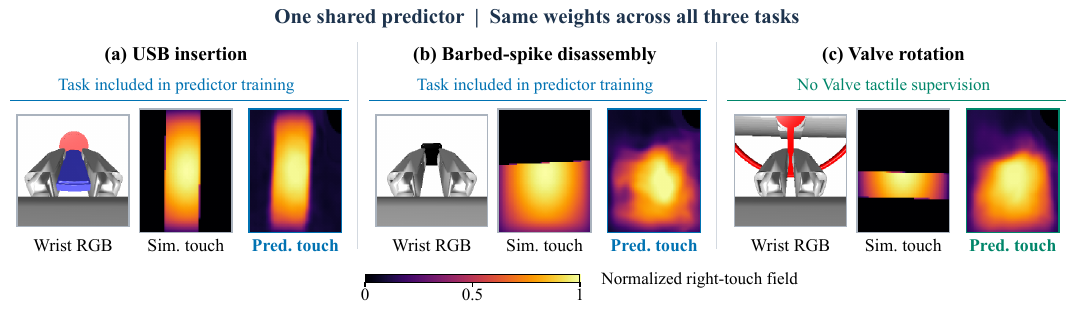}
  \caption{One shared predictor supplies current-touch estimates across
  simulation tasks. Each group pairs wrist RGB, simulator right-touch, and
  predicted right-touch on a common $[0,1]$ scale. USB and Barbed-spike are
  included in its six-task training; Valve provides no tactile supervision
  to this predictor. All predictions use the same weights and no task-varying
  token. Cases follow the fixed median-contact episode rule, independent of
  prediction error and control outcomes; visible shape and extent differences
  illustrate the limits of reconstruction.}
  \label{fig:task_predictions}
\end{figure*}

\subsection{Policy learning and execution with predicted touch}
In condition labels, the term before the slash gives the tactile input during
policy training, and the term after it gives the input during execution.
\emph{Simulator touch} (T/T) uses simulator-provided touch at both stages;
\emph{predicted touch} (P/P) uses predicted touch at both stages. The predictor
is trained with tactile supervision and then frozen before policy training.
Thus, P/P describes the policy's inputs, not the predictor's supervision.

A separate \emph{fixed-policy input-replacement} study supplies zero, simulator
touch, Pretrained, task-adapted predictions, or shared predictions to an
already trained tactile policy. It tests each input source with the policy
held fixed.

\Needspace{3\baselineskip}
Its checkpoints and evaluation results remain separate from the
main policy-learning comparison.

Our simulation experiments use ContactWorld's environments and downstream
world-model policy~\cite{zhang2026contactworld}. PredTac supplies the tactile
observation channel to this controller, whose visual--tactile encoders,
latent dynamics model, and cross-entropy-method (CEM) planner follow
ContactWorld. CEM uses horizon 6, 100 candidates, 8 elites, and 4 iterations.

The visual system is a separately trained RGB-only controller with no tactile
input. The predicted-touch system additionally uses an external predictor
that consumes visual observations and robot state.

The main predicted-touch comparison uses predictors trained from scratch: USB includes
the temporal and geometry losses, while Barbed-spike uses a right-touch
predictor that is also reused on Valve without Valve tactile supervision.
They share the three-frame architecture with the source checkpoint but have
independently learned weights. The auxiliary fine-tuning and shared-predictor
studies use separate checkpoints and do not replace these main policy-learning results.

\RealMethodExtension

\section{Experiments}
\begin{table*}[t]
\begin{LayoutInlineTable}
  \caption{Success rates and differences across controller conditions in
  simulation.}
  \label{tab:systems}
  \centering
  \footnotesize
  \setlength{\tabcolsep}{2.1pt}
  \renewcommand{\arraystretch}{1.08}
  \begin{tabularx}{\textwidth}{@{}L{1.8cm}*{4}{Y}*{3}{C{2.4cm}}@{}}
    \toprule
    & \multicolumn{4}{c}{Success (\%)} &
    \multicolumn{3}{c}{Difference (pp)} \\
    \cmidrule(lr){2-5}\cmidrule(lr){6-8}
    Task & \shortstack{Simulator touch\\(T/T)} &
    \shortstack{Predicted touch\\(P/P)} &
    \shortstack{Touch removed\\(P/0$_b$)} & Vision & P/P$-$P/0$_b$ &
    P/P$-$Vision & T/T$-$P/P \\
    \midrule
    USB & 27.0 & \textbf{27.0} & 15.3 & 19.0 & +11.7 &
    \textbf{+8.0} & 0.0 \\
    Barbed-spike & 52.3 & \textbf{52.0} & 50.3 & 43.3 & +1.7 &
    \textbf{+8.7} & +0.3 \\
    Valve & 48.7 & \textbf{44.7} & 36.3 & 31.0 & +8.3 &
    \textbf{+13.7} & +4.0 \\
    \bottomrule
  \end{tabularx}
  \vspace{0.4ex}

  \raggedright\footnotesize Entries report terminal-joint success (\%) from
  300 rollouts per condition under the goal-offset protocol; pp denotes
  percentage points. Labels before/after the slash denote inputs during policy
  training/execution. Touch removed (P/0$_b$) keeps the predicted-touch policy
  but zeroes current and goal touch and skips prediction; $b$ denotes broad
  removal. Zero current (Table~\ref{tab:current_only}) zeroes only current touch.
  Vision is a separately trained RGB-only controller. Bold highlights P/P
  success and its gains over Vision; it does not indicate significance.
\end{LayoutInlineTable}
\end{table*}

\begin{table*}[t]
\begin{LayoutInlineTable}
  \caption{Policy success with current-touch input interventions in simulation.}
  \label{tab:current_only}
  \centering
  \footnotesize
  \setlength{\tabcolsep}{2.5pt}
  \renewcommand{\arraystretch}{1.07}
  \begin{tabularx}{\textwidth}{@{}l*{3}{Y}@{}}
    \toprule
    & \multicolumn{3}{c}{Success (\%) / drop from Correct (pp)} \\
    \cmidrule(lr){2-4}
    Condition & USB & Barbed-spike & Valve \\
    \midrule
    Correct prediction & 27.7 / -- & 51.0 / -- & 44.0 / -- \\
    Zero current & 12.7 / \textbf{15.0} & 45.7 / 5.3 & 37.0 / 7.0 \\
    Random donor & 23.7 / 4.0 & 42.0 / \textbf{9.0} & 40.3 / 3.7 \\
    Contact matched & 23.3 / 4.3 & 47.0 / 4.0 & 42.3 / 1.7 \\
    Contact + intensity & 25.7 / 2.0 & 48.0 / 3.0 & 43.7 / 0.3 \\
    Local temporal & 25.7 / 2.0 & 49.3 / 1.7 & 43.7 / 0.3 \\
    Global temporal & 27.0 / 0.7 & 48.0 / 3.0 & 41.7 / 2.3 \\
    Spatial histogram & 24.0 / 3.7 & 45.3 / 5.7 & 33.3 / \textbf{10.7} \\
    \bottomrule
  \end{tabularx}
  \vspace{0.4ex}

  \raggedright\footnotesize Bold marks the largest observed drop per task,
  not statistical significance. Zero current is prespecified; other
  corruptions are exploratory (definitions in
Section~\ref{sec:current_interventions}). Absolute rates are not
  comparable to Table~\ref{tab:systems}.
  USB/Barbed-spike use Final-or-Initial Joint; Valve uses Final Joint. The suite
  contains 72 runs and 7,200 rollouts from 155, 155, and 95 source episodes.
\end{LayoutInlineTable}
\end{table*}

\RealResultsDisplay

\subsection{Hardware and computing platform}
The real-robot setup comprises a six-axis RealMan RM65B arm, a WHEELTEC
gripper, and two PaXini M3025 tactile arrays, one on each finger. Two Orbbec
Femto Bolt cameras provide wrist RGB for policy input and an external RGB
view for recording. Real-robot model training uses a server equipped with
NVIDIA GeForce RTX 3090 GPUs. Online tactile prediction and ACT inference
run in PyTorch on a local Windows computer with an NVIDIA GeForce RTX 4060
Laptop GPU (8\,GB), which also handles robot communication and control.

\subsection{Tasks and goal-offset evaluation protocol}
Legacy USB, Barbed-spike, and Valve trials restore a demonstration state and a
goal 24/12/36 recorded steps ahead, with at most 30/15/45 planning steps.
Terminal success requires the final keypoint positions, end-effector
pose, and joint configuration to satisfy their respective thresholds. It measures a local subtrajectory,
not standard-reset semantic completion; Barbed-spike and Valve retain recorded
gripper schedules.

The current-only suite uses Final Joint on Valve and Final-or-Initial Joint on
USB/Barbed-spike; ``Initial'' means the restored start already meets the goal,
not later intermediate success. Its effects are compared only within-suite and
absolute rates are not pooled with the four-system results.

\subsection{Systems, interventions, and seeds}
We compare policies trained and executed with simulator touch (T/T),
predicted touch (P/P), or visual inputs (Vision). Vision is a separately
trained visual system without tactile inputs.

\emph{Touch removed} (P/0$_b$) reuses the trained predicted-touch policy without
retraining. During execution, both current and goal touch are set to zero
and predictor execution is skipped; $b$ denotes broad removal of these
tactile inputs. By contrast, \emph{Zero current} in
Table~\ref{tab:current_only} sets only current touch to zero, retaining
predicted goal touch and the predictor. Tables~\ref{tab:systems} and
\ref{tab:current_only} use different protocols and are analyzed separately.

Each condition uses three rollout/CEM strata of 100 environments, with the
same evaluation cases across the four systems within each stratum. The
predicted-touch and visual policies each use one training seed (42 for
predicted touch); the simulator-touch policy's training seed is unrecorded.

\Needspace{8\baselineskip}
\subsection{Current-touch interventions}
\label{sec:current_interventions}
Table~\ref{tab:current_only} holds policy and predictor weights fixed and
retains goal touch. \emph{Correct prediction} leaves current touch unchanged;
\emph{Zero current} sets it to zero.

Donor conditions permute current predictions across parallel rollouts from
different source episodes using a fixed one-to-one assignment.
\emph{Random donor} imposes no contact-matching constraint.
\emph{Contact matched} minimizes disagreement in contact/no-contact sequences
from the reference rollouts. \emph{Contact + intensity} additionally matches
predicted mean and peak magnitude and active area, with contact agreement
taking priority. These approximate matches do not preserve exact contact shape.

\emph{Local temporal} substitutes a prediction from one or two decisions
earlier in the same rollout. \emph{Global temporal} replays historical blocks
selected 4--12 decisions in the past, using available history. \emph{Spatial
histogram} reorders spatial blocks within the current field while preserving
its exact pixel-value histogram.

\subsection{Evaluation metrics}
Task performance is measured by success rate and percentage-point differences.
Real-robot entries include successes/trials; task means use equal weights.
Comparisons are descriptive and do not establish measured-touch equivalence
or retraining robustness.

Prediction quality is summarized by full-field MAE (lower is better) and
contact-mask IoU at a threshold of 0.05 (higher is better).

\raggedbottom
\RealExperimentExtension

\subsection{Auxiliary predictor evaluation}
USB reconstruction is evaluated on 20 episodes at the policy's tactile
resolution. Valve adaptation compares fine-tuning and training from scratch
with 8, 20, or 82 target-training episodes, 500 updates, one predictor seed,
and the same 10 evaluation episodes. It replaces inputs to one fixed tactile
policy rather than retraining that policy. The six-task shared-predictor
study likewise holds each task's tactile policy fixed and compares source
and shared predictions. These two auxiliary studies use 300 rollouts per
task and condition; reconstruction and control are evaluated separately.

For transfer to Valve, we replace the original predictor of the P/P policy
in Table~\ref{tab:systems} with the shared predictor, keeping the policy
weights fixed. The shared predictor is fine-tuned on six other simulation
tasks and uses no Valve tactile data for that fine-tuning.

\section{Results}
\subsection{Task performance with predicted touch}
In the simulation goal-offset evaluations, predicted-touch policies (P/P)
achieve 27.0\%, 52.0\%, and 44.7\% success on USB, Barbed-spike, and Valve
(Table~\ref{tab:systems}), exceeding the separately trained visual system by
8.0--13.7 percentage points. Policies using simulator touch (T/T) achieve 27.0\%,
52.3\%, and 48.7\%, respectively. These observed rates are close but do not
establish equivalence. The differences from Vision compare separately trained
controllers under the goal-offset protocol and do not isolate the effect of
tactile content. The interventions below examine the role of tactile content
within a fixed policy.

\Needspace{7\baselineskip}
\subsection{Does the policy use predicted touch?}
With the predicted-touch policy fixed, removing current predicted touch reduces success
by 15.0, 5.3, and 7.0 points on USB, Barbed-spike, and Valve
(Table~\ref{tab:current_only}). Removal shows input sensitivity; the replacement
conditions further test which aspects of the contact content matter.
On Barbed-spike, the drop is 9.0 points with random donors and 3.0 points
when donors match contact and intensity summaries. On Valve, spatial-block
reordering reduces success by 10.7 points despite preserving the value
histogram, suggesting sensitivity to spatial arrangement beyond aggregate
intensity. Temporal replacements yield smaller observed drops, so this suite
does not establish precise temporal dependence.

The observed drops vary across tasks and replacement types, indicating that
policies differ in their sensitivity to contact content. These replacement
comparisons are exploratory. \emph{Touch removed} in Table~\ref{tab:systems}
also removes goal touch and skips prediction, with observed drops of 11.7,
1.7, and 8.3 points on USB, Barbed-spike, and Valve. Its intervention and
evaluation protocol differ from those in Table~\ref{tab:current_only}, so the
two sets of results are interpreted separately.

\RealResultsExtension

\subsection{Prediction quality and reuse}
The best USB reconstruction in the policy-resized evaluation is obtained
with structured target fine-tuning: MAE is 0.0282 and contact-mask IoU is
0.8845. These metrics characterize the adapted predictor; the task-policy
comparison in Table~\ref{tab:systems} retains its declared checkpoints.
In the separate Valve input-replacement study, target-supervised predictors
raise observed success from 37.3\% for the source predictor to as high as
53.3\% when trained from scratch with 82 target-training episodes. Both
fine-tuning and training from scratch improve success; fine-tuning does not
outperform scratch at the matched budgets.

Figure~\ref{fig:task_predictions} pairs simulator touch and predictions from
one shared predictor on USB and Barbed-spike, which contribute predictor
training data, and on Valve, which contributes no tactile labels to this
predictor. The same weights are used without a task-varying input token.
In fixed-policy input replacement on the six training tasks, the shared
predictor increases observed mean success from 41.7\% to 43.7\%; five tasks
improve, while Lidded-loose decreases from 29.7\% to 24.0\%.
The examples retain visible differences in contact shape and extent,
particularly on Valve.

On Valve, the P/P policy achieves 44.7\% success with its original predictor
(Table~\ref{tab:systems}) and 44.0\% after replacing it with the shared
predictor, without retraining the policy. This 0.7-percentage-point decrease
indicates that the shared predictor can be reused by the same Valve policy
with a small observed change in success, despite receiving no Valve tactile
data during its six-task fine-tuning.

\section{Discussion and Limitations}
The results support using predicted touch during policy learning and execution.
Task comparisons measure the performance of the resulting systems, while
fixed-policy interventions examine their use of tactile content. The latter
indicate sensitivity to different contact properties across tasks: matching
contact and intensity reduces the donor-replacement drop on Barbed-spike,
whereas rearranging spatial blocks reduces Valve performance despite preserving
the value histogram. These observations do not imply that every reduction in
reconstruction error improves control.

Predictor reuse also has limits. The shared predictor improves observed
performance on five of the six training tasks but reduces success on
Lidded-loose. Thus, reusing a predictor does not consistently improve the
performance of an existing tactile policy across tasks.

The experiments do not isolate the contributions of state input, temporal
fusion, contact conditioning, or decoding. The target-training comparisons
vary the objective or initialization, so they do not establish the necessity
of individual network components.
\RealDiscussionBoundary

\RealDiscussionExtension

\section{Conclusion}
\method{} uses predicted current touch as an explicit input during policy
learning and execution. Once trained with tactile supervision, the frozen
predictor allows the policy to learn and act with generated tactile inputs.
In simulation goal-offset evaluations, predicted-touch controllers improve
on the visual system, and fixed-policy interventions indicate task-dependent
sensitivity to contact content, including its spatial arrangement on Valve.
\RealConclusionExtension

Together, task performance and input interventions support predicted touch
as a useful control representation and help characterize how policies respond
to its contact content. Evaluation across new task conditions and repeated
policy training remains necessary to establish generalization and training
robustness.

\section*{Acknowledgment}
OpenAI Codex assisted with manuscript editing and code used to prepare
Figs.~1--4. The authors reviewed and edited the resulting text and figures
and take responsibility for the final manuscript.

\IEEEtriggeratref{16}
\bibliographystyle{IEEEtran}
\bibliography{references}
\end{document}

%% file: real_extension_en.tex
\input{real_success_slots_en.tex}

\makeatletter

\newenvironment{RealInlineTable}{\begingroup\def\@captype{table}}{\endgroup}
\makeatother

\renewcommand{\RealMethodExtension}{%
\subsection{Real-robot policy learning and execution}
On all three tasks, visual ACT~\cite{zhao2023learning} uses current wrist
RGB and robot state; measured-touch ACT adds measured touch; predicted-touch
ACT adds predictions during both policy training and execution. Predictors
learn current touch from causal RGB/state histories with measured-touch
supervision, then remain frozen. Measured touch is not an input to the
predicted-touch policy.

Opposing PaXini M3025 arrays provide
$X_t\in\mathbb{R}^{2\times77\times3}$: two fingers, 77 taxels each,
and $(F_x,F_y,F_z)$, with normal force $F_z$. Predictions preserve this ordering.

ACT uses $128\times128$ wrist images, eight-step action chunks, and
10\,Hz control, retaining training-time preprocessing, action interfaces,
and tactile normalization. External RGB records execution.}

\renewcommand{\RealExperimentExtension}{%
\Needspace{8\baselineskip}
\subsection{Real-robot data and task protocol}
Paired demonstrations supply tactile targets for predictor supervision
and actions for policy learning. USB and Valve have 49 (39/5/5) and
39 (29/5/5) demonstrations, respectively, with episode-level
train/validation/test counts in parentheses. The tactile-conditioned
Barbed policies use 75 (60/12/3), excluding five training episodes from
80 without changing the frozen predictor split. Barbed preprocessing
compacts waiting, repairs gripper targets, and preserves causal histories;
execution suppresses upward commands during alignment.

Barbed visual ACT uses 512 hidden units and seven decoder layers with
10,000 training updates; its tactile ACTs use 256 units and four decoder
layers with 4,000 refinement updates. Thus Barbed compares deployed
policies with differing model and training configurations.

USB inserts a pregrasped connector with fixed orientation. Barbed comprises
grasping, lateral disengagement, extraction, and holding without release.
Completion is judged manually for both tasks. Valve starts ungrasped: the
policy grasps the valve, rotates it clockwise toward $90^\circ$, and
maintains its grasp without autonomous release. Valve success is defined by
the recorded rotation angle $\theta$ satisfying
$85^\circ\leq\theta\leq95^\circ$.

Each task/ACT condition has 30 physical evaluation trials, totaling 270
trials separate from the demonstrations. Table~\ref{tab:real_success}
reports successes/trials and rates; the task mean uses equal weights.}

\renewcommand{\RealResultsDisplay}{%
\begin{figure*}[t]
  \centering
  \includegraphics[width=\textwidth]{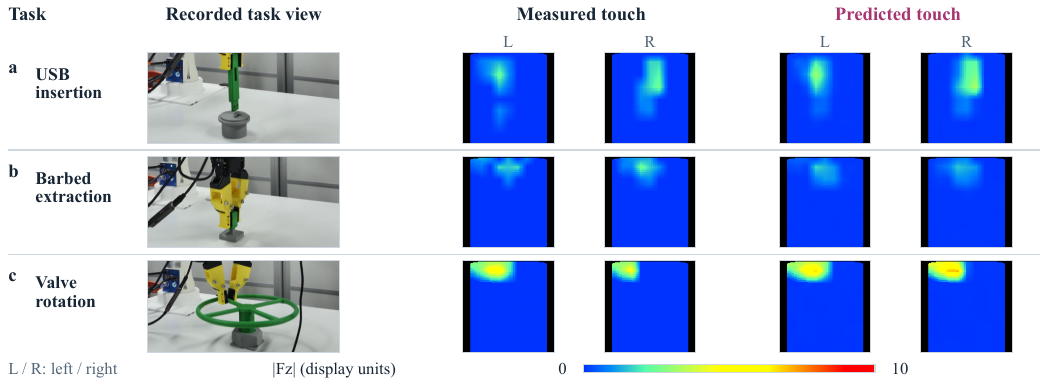}
  \caption{Recorded real-robot contact across three tasks. Each row pairs
  external RGB and measured/predicted left/right (L/R) $|F_z|$ from one inference
  record. Samples are nearest the median measured total $|F_z|$ among valid
  motion-enabled observations, selected without prediction error. Maps share
  the runtime projection and fixed 0--10 vendor-decoded force scale.
  Examples do not estimate success.}
  \label{fig:real_examples}
\end{figure*}

\begin{table*}[t]
\begin{RealInlineTable}
  \centering
  \caption{Real-robot task success with visual, measured-touch, and
  predicted-touch ACT (30 trials per task and condition).}
  \label{tab:real_success}
  \footnotesize
  \renewcommand{\arraystretch}{1.0}
  \begin{tabularx}{\textwidth}{@{}L{4.4cm}YYY@{}}
    \toprule
    Model / condition & USB & Barbed & Valve \\
    \midrule
    Visual ACT & \RealUSBVisual & \RealBarbedVisual & \RealValveVisual \\
    Measured-touch ACT & \RealUSBMeasured & \RealBarbedMeasured & \RealValveMeasured \\
    \textbf{Predicted-touch ACT} & \RealUSBPredicted & \RealBarbedPredicted & \RealValvePredicted \\
    \bottomrule
  \end{tabularx}\par
  \vspace{3pt}
  \noindent\parbox[t]{\textwidth}{\footnotesize Entries: success rate (successes/trials).
  USB/Barbed completion is judged manually; Valve uses a recorded rotation
  of $85^\circ$--$95^\circ$, with no autonomous release.
  Differences between conditions are descriptive. Bold highlights
  predicted-touch ACT, not necessarily the highest rate.}
\end{RealInlineTable}
\end{table*}
}

\renewcommand{\RealResultsExtension}{%
\Needspace{5\baselineskip}
\subsection{Real-robot task success with predicted touch}
Predicted-touch ACT improves on visual ACT across all three real-robot tasks.
It achieves 70.0\%, 50.0\%, and 90.0\% success on USB, Barbed, and Valve
(Table~\ref{tab:real_success}), with gains of 66.7, 26.7, and 53.3 percentage
points. Its observed differences from measured-touch ACT are $-6.7$, $-3.3$,
and $+3.3$ points. The three-task means are 70.0\% with predicted touch,
72.2\% with measured touch, and 21.1\% with visual inputs.

These observations show that policies trained and executed with predicted
touch approach the measured-touch condition on the evaluated tasks. The
comparisons are descriptive and do not establish equivalence.
Figure~\ref{fig:real_examples} pairs recorded measured and predicted force
maps for each task.}

\renewcommand{\RealDiscussionBoundary}{The simulation results in
Tables~\ref{tab:systems}--\ref{tab:current_only} use fixed checkpoints and
goal-offset evaluations. They are analyzed separately from the full physical
task trials in Table~\ref{tab:real_success}. The comparisons do not assess
variability across repeated policy training.}

\renewcommand{\RealDiscussionExtension}{}

\renewcommand{\RealConclusionExtension}{Across the three real-robot tasks,
policies trained and executed with predicted touch improve on visual ACT and
approach observed measured-touch performance, without measured touch as a
policy input.}

%% file: real_success_slots_en.tex
\newcommand{\RealUSBVisual}{3.3\% (1/30)}
\newcommand{\RealUSBMeasured}{76.7\% (23/30)}
\newcommand{\RealUSBPredicted}{\textbf{70.0\%} (21/30)}
\newcommand{\RealBarbedVisual}{23.3\% (7/30)}
\newcommand{\RealBarbedMeasured}{53.3\% (16/30)}
\newcommand{\RealBarbedPredicted}{\textbf{50.0\%} (15/30)}
\newcommand{\RealValveVisual}{36.7\% (11/30)}
\newcommand{\RealValveMeasured}{86.7\% (26/30)}
\newcommand{\RealValvePredicted}{\textbf{90.0\%} (27/30)}